\documentclass[sigconf]{acmart}
\usepackage{booktabs} 
\usepackage{subfigure}
\usepackage{blindtext}
\usepackage{graphicx}
\usepackage{csquotes}
\usepackage{caption}
\usepackage{amsmath}
\usepackage{siunitx}
\usepackage{multicol}

\usepackage{xcolor,colortbl}
\definecolor{Gray}{gray}{0.90}
\newcolumntype{a}{>{\columncolor{Gray}}c}
\usepackage{xcolor,soul}
\definecolor{light-gray}{gray}{0.95}
\sethlcolor{light-gray}

\graphicspath{{Figures/}}
\newcommand{\fig}[1]{Fig.~\ref{#1}}
\newcommand{\tb}[1]{Table~\ref{#1}}
\newcommand{\eq}[1]{(\ref{#1})}

\usepackage{enumitem,booktabs}
\usepackage[referable]{threeparttablex}
\renewlist{tablenotes}{enumerate}{1}
\makeatletter
\setlist[tablenotes]{label=\tnote{\alph*},ref=\alph*,itemsep=\z@,topsep=\z@skip,partopsep=\z@skip,parsep=\z@,itemindent=\z@,labelsep=.2em,leftmargin=*,align=left,before={\footnotesize}}
\makeatother

\copyrightyear{2018}

\begin{document}
\title{Voltage Mode Winner-Take-All Circuit for Neuromorphic Systems}


\author{Abdullah M. Zyarah$^{\dagger}$$^{\S}$ and Dhireesha Kudithipudi$^{\dagger}$}
\orcid{0000-0001-8220-528}
\affiliation{
\institution{$^{\S}$Department of Electrical Engineering, University of Baghdad, Iraq \\ 
$^{\dagger}$NuAI Lab, Department of Electrical and Computer Engineering, University of Texas at San Antonio}
}
\email{abdullah.zyarah@uob.edu.iq}


\begin{abstract}
Recent advances in neuromorphic computing demonstrate on-device learning capabilities with low power consumption. One of the key learning units in these systems is the winner-take-all circuit. In this research, we propose a winner-take-all circuit that can be configured to achieve k-winner and hysteresis properties, simulated in IBM 65 nm node. The circuit dissipated 34.9 $\mu$W of power with a latency of 10.4 ns, while processing 1000 inputs. The utility of the circuit is demonstrated for spatial filtering and classification. 
\end{abstract}

%
%

\ccsdesc[300]{Computing methodologies~Neural networks}
\ccsdesc[300]{Computer systems organization~Analog computers}

\keywords{Winner-take-all, softmax, Neuromorphic circuits}

\maketitle

\section{Introduction}
Enabling real-time processing and low power consumption in neuromorphic chips is a real challenge with the unprecedented increase in the amount of data that need to be processed.  A large number of these chips implement self-organizing neural networks and use competitive learning, which is based on selecting a winning neuron among several others. Using conventional daisy chain methods in finding the winning class is a performance bottleneck for learning and classifying ~\cite{wu2017low}. Thus, winner-take-all (WTA) circuit is usually recommended.
 
Winner-take-all (WTA) circuits are fundamental units that dominate most neural network architectures especially in classification applications. The WTA circuit performs local competitions among a set of competing inputs to select one winning output node. While the output node that is associated with the largest input (voltage/current) emerges as a winner, all other nodes are suppressed. The WTA has been used as a key component in a large variety of applications such as rank modulation coding of multi-level-cell memory~\cite{kim2016rank}, extracting the salient features in images~\cite{mahowald1989cooperative}, vector quantization processors~\cite{ogawa2002general}, and analog classifiers~\cite{ramakrishnan2014vector}. Although most of the WTA circuits that are proposed in literature are analog circuits, there were some attempts to implement digital WTA circuits. While the digital WTA circuits offer more stability and precision in comparison to its analog counterpart~\cite{yoon2015vlsi}, the analog WTAs still provide more compact and energy efficient circuits~\cite{rahman2009high}. Moreover, the analog WTA architecture is not affected by bit precision as in digital. This makes it well suited for neural network architectures when deploying in mobile and energy constrained platforms. 

Several WTA circuits have been proposed in literature~\cite{lazzaro1989winner, choi1993high, prommee2011cmos,indiveri2001current,ramakrishnan2014vector, waugh1993analog, urahama1995k,fish2001cmos, yoon2015vlsi, starzyk1996voltage, kulej2017sub, fish2005high}. One of the widely used WTA circuits is proposed by Lazzaro et al. in 1989~\cite{lazzaro1989winner}. The circuit is compact and energy efficient but it has limited precision and operation speed. A variant of Lazzaro circuit is proposed by Indiveri et al. in 2001 and Ramakrishnan et al. in 2014. The former endows the circuit hysteresis, lateral inhibition and excitation properties~\cite{indiveri2001current}, while the latter enables multiple winners~\cite{ramakrishnan2014vector}. In 1993, Waugh et al. proposed a WTA circuit in which neurons compete locally via a constraint applied through Kirchoff's current law (KCL)~\cite{waugh1993analog}, the resultant circuit is very simple as one transistor is used to model each cell in the WTA circuit. However, the circuit requires additional components to sense its output.  Then, Urahama et al., in 1995, presented a WTA circuit that can handle input voltage/current. However, the circuit resolution is limited to about 0.2v, which can be reduced by increasing network complexity~\cite{urahama1995k}. In 2015, Yoon et al. proposed a simplified digital WTA circuit~\cite{yoon2015vlsi}. The circuit can be scalable to hundreds of inputs with graceful degradation in circuit latency, but it would not be preferable when it comes to energy constrained platforms. Sub 0.5v bulk driven WTA circuit that is based on a voltage follower is proposed by Kulej et al. in 2017~\cite{kulej2017sub}. The circuit works on voltage level as low as 0.3v and relays the maximum input voltage to one shared output node rather than its index. 

\begin{figure*}[h!tb]
\begin{center}
\includegraphics[width = 0.9 \textwidth]{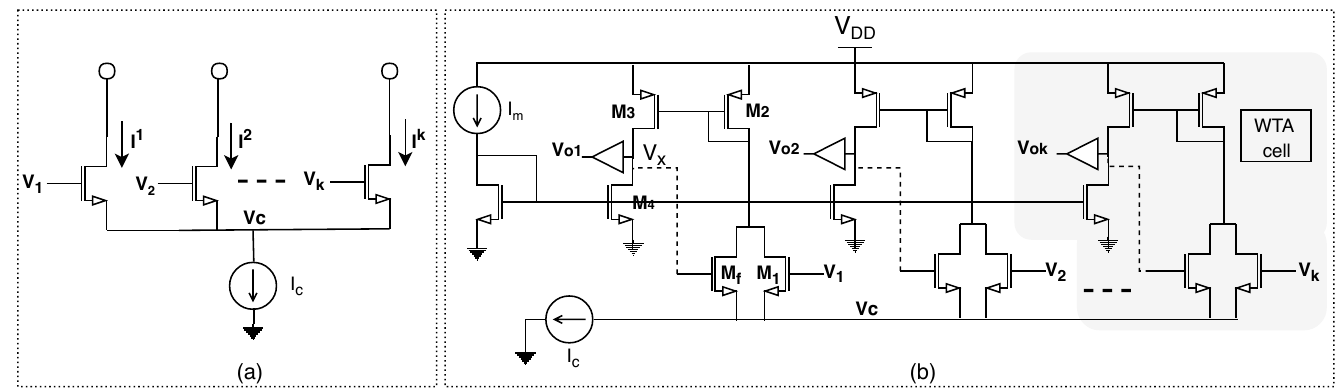}
\caption{(a) Current conveyor circuit proposed by~\cite{andreou1991current} to implement a WTA function. (b) The proposed WTA circuit ($k$ number of cells) with local excitatory feedback.}
\label{wta_cct}
\end{center}
\end{figure*}

In this research, a WTA circuit that operates at high speed and dissipates low energy is proposed for neuromorphic applications. Unlike some of the circuits in literature, the proposed circuit is a voltage-mode so that it does not require any conversion circuitry when interfaced with other signal processing IC~\cite{choi1993high}. Another potential advantage of using voltage-mode is that the voltage can be stored in a capacitive storage whereas sustaining the information in current-mode requires a continuous supply of current leading to more energy dissipation~\cite{starzyk1996voltage}. The proposed circuit is built-up from the current conveyor circuit ~\cite{andreou1991current}. Key features of the proposed circuit include: 
\begin{itemize}
\item Asynchronous operation and parallel input processing. The circuit also offers high drive strength achieved via buffering all outputs.
\item Configurable sensitivity level to account for signal variations and noise. Furthermore, the circuit allows multiple winners rather than only one winner. Expands the applicability to different neural algorithms for pattern matching.
\item The proposed circuit implements a softmax function to suppress the current in losing cells while keeping it maximum in the winners. Consequently, low power consumption is achieved even when the circuit scales up to
hundreds of cells.
\end{itemize}



\section{WTA Circuit Description}
The proposed WTA circuit is a variant of the circuit proposed in~\cite{andreou1991current}, which is shown in \fig{wta_cct}-(a). The circuit models a simple local competitive algorithm which is naturally imposed through KCL~\cite{elfadel1994softmax}. Each branch in the circuit has one NMOS transistor to capture the input signal of one competitor. The competitors interact with each other through the shared point $V_c$. When inputs are presented to the circuit, the potential of $V_c$ follows the input with highest voltage and turns off all the other transistors. The cell conveying most of the bias current, $I_c$,  is identified as a winner. Given that all the transistors operate in subthreshold regime, applying an input voltage $V_{G}$ at the gate of the transistor in branch $i$ results in a current $I$, which can be approximated by \eq{subth_cur}~\cite{razavi2017design}:

\begin{equation}
I^{i} = I_o~(\frac{W}{L})~e^{\frac{V^i_{GS}}{nU_T}}
\label{subth_cur}
\end{equation}

\noindent where $I_o$ is the zero-bias current for the given device, $\frac{W}{L}$ is the transistor channel width to length ratio,  $U_T$ and $n$ indicate the thermal voltage and the subthreshold slope coefficient, respectively. For the given circuit with $k$ branches, according to KCL, the branches' current should sum up to $I_c$, as given by \eq{sum_current}. By using \eq{subth_cur} and \eq{sum_current}, we can solve for the current flowing in each branch as shown in \eq{branch_current}, which is identical to the softmax equation. 

\begin{equation}
I_c = \sum^k_{i=1} I^i
\label{sum_current}
\end{equation}

\begin{equation}
I^i = {I_c}~\frac{e^{\frac{V^i_{G}}{nU_T}}}{\sum\limits^k_{i=1} e^{\frac{V^i_{G}}{nU_T}}}
\label{branch_current}
\end{equation}

According to \eq{branch_current} the output of the circuit is encoded in the form of a current which is not definite and requires a special converter when integrated with digital processors. Thus, this work emphasizes more on the voltage mode WTA. For the circuit in \fig{wta_cct}-(a), a naive way of modifying the circuit to achieve the output in the form of a voltage is by adding an active load using a diode connected PMOS transistor and capturing the voltage at the NMOS transistor drain. The downside of this approach is that the input difference at the circuit inputs will be approximately the same as the output, i.e. $V^i_{G} - V^j_{G} \approx V^i_{DS} - V^j_{DS}$, where $i$ and $j$ are the indexes of two different branches. Furthermore, this circuit will have low resolution and precision, and requires continuous tweaking as the circuit size changes to ensure proper circuit operation. Therefore, in order to maintain the same normalized exponential relationship between the input and output, the current in each branch is sent to a current comparator via a current mirror formed by $M_2$ and $M_3$, as shown in \fig{wta_cct}-(b). The mirrored current is compared to a fixed reference current resulting in a voltage drop across the point $V_x$, which can be calculated using~\eq{vx_eq}:

\begin{equation}
V^i_{x} = \frac{1}{\lambda_4} \Big[\frac{2g~I_{2}}{\beta_4 (V_{GS4} - V_{th4})^2} - 1 \Big]
\label{vx_eq}
\end{equation}

\noindent where $\lambda$ is the channel-length modulation, $\beta$ is the transconductance parameter, $g$ and $V_{th}$ denote the current mirror gain between $M_2$ and $M_3$ and transistor threshold voltage. By substituting \eq{branch_current} in \eq{vx_eq}, the output node, $V_x$, is calculated in \eq{vout_eq}: 

\begin{equation}
V^i_{x} =  \xi_4{I_c}~\frac{e^{\frac{V^i_{G4}}{nU_T}}}{\sum\limits^k_{i=1} e^{\frac{V^i_{G}}{nU_T}}} - \frac{1}{\lambda_4} 
\label{vout_eq}
\end{equation}

\begin{figure*}[h!tb]
\centering
\subfigure[]{\includegraphics[width=50mm, height=40mm]{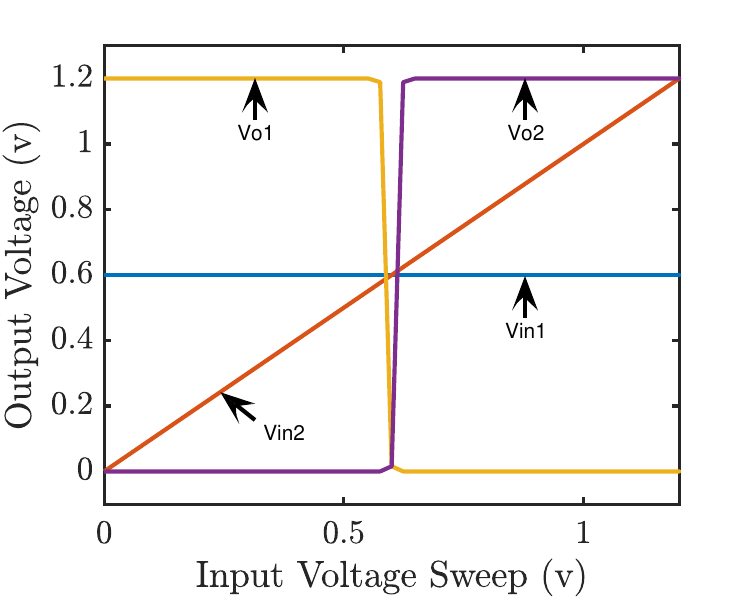}}
\subfigure[]{\includegraphics[width=50mm, height=40mm]{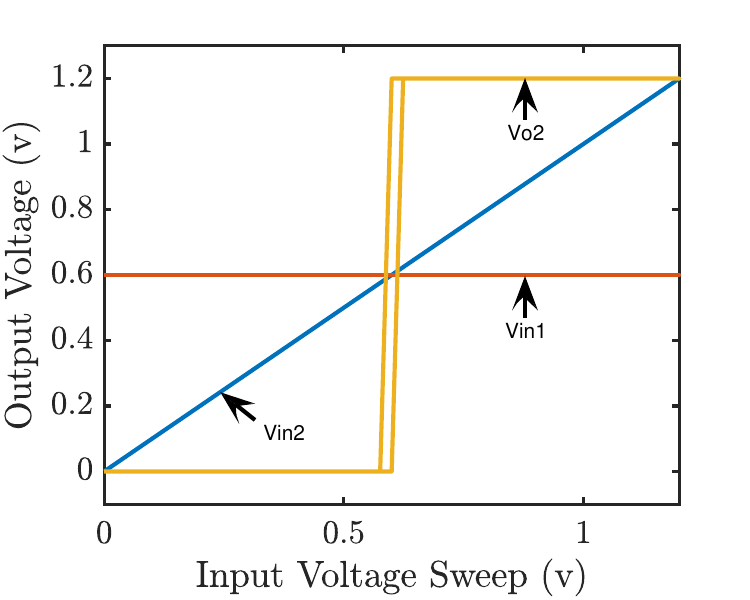}}
\subfigure[]{\includegraphics[width=50mm, height=40mm]{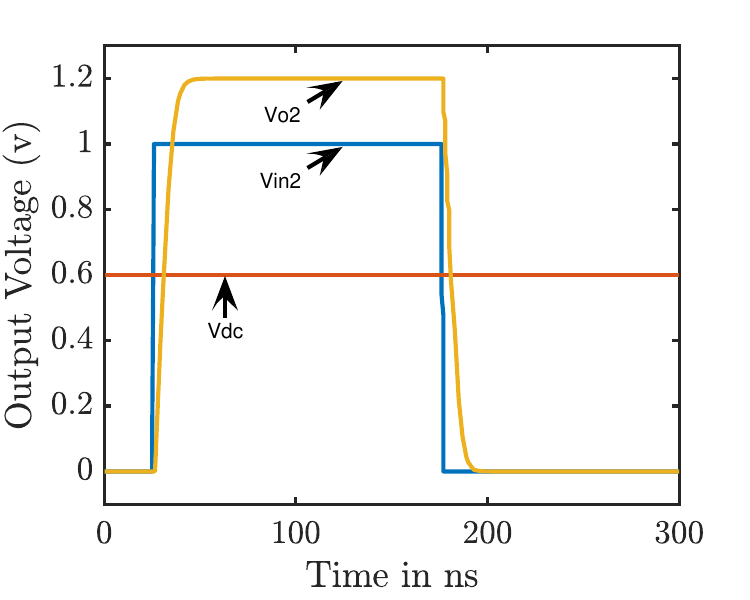}}
\caption{(a) The illustration of 2-cells WTA circuit operation to measure the minimum voltage required to cause cells output status to flip. (b) Hysteresis characteristics of the proposed circuit when adding the local excitatory feedback. (c) Transient response while driving 1 pF capacitive load.}
\label{accur}
\end{figure*}

Due to the fact that $\xi_4$ is approximately constant and is given by $\frac{2g}{\lambda_4\beta_4 (V_{GS4} - V_{th4})^2}$, \eq{vout_eq} indicates that the output voltage $V^i_x$ for branch $i$ has a normalized exponential relationship with the input $V^i_G$. Such relation has a unique benefit for WTA circuit because it maximizes the difference between the inputs. It generously rewards the input with the highest value and punishes the losing ones. Furthermore, this mechanism makes the power consumption of the circuit extremely low. Most of the power consumption is dominated by the winning cells which are low in number compared to losing cells.
It is important to notice here that unlike most other WTA circuits, discussed in literature, all outputs are buffered to provide enough driving capabilities when transmitting signal across long distances. Also, few of the previous WTA circuits are endowed with a hysteresis mechanism to increase network stability and prevents the selection of a potential winner unless they are strong enough. Due to the fact that the hysteresis is achieved via a local excitatory feedback, some of these circuits require a reset process to any competition as in~\cite{fish2005high}. In the proposed WTA circuit, the hysteresis characteristics\footnote{It is recommended to use dual voltage source when the WTA circuit with the hysteresis characteristics is used. This is to increase the input voltage range sensed by the circuit.} is introduced via the positive feedback formed by the transistor $M_f$. The width of the hysteresis curve is modulated by the transistors ($M_1$ and $M_f$) size ratio. Additionally, having a current comparator improve the stability further as it imposes a threshold current that needs to be crossed to switch cells status. The other advantage of using the current comparator is that it enables more than one winner, which is a desirable feature especially in pattern matching applications. 





\section{Analysis and Design Methodology} 
To effectively investigate the characteristics of the proposed WTA circuit, the circuit is simulated under different conditions and setups. Initially, the circuit is simulated to estimate the minimum voltage difference between two cells to cause a state flip at the output. This is done by using two cells only. One of the cells is tied to a fixed voltage of 0.6v while the other cell input is swept between 0 and 1.2v. It is found that at least a 4mv difference is required to flip the outputs. \fig{accur}-(a) illustrates the circuit output while sweeping cell-2 input voltage and fixing cell-1 input. The same experiment is repeated in the presence of local excitatory feedback (positive feedback) but with hysteresis sweep. This is to capture the hysteresis characteristics curve, which is shown in \fig{accur}-(b).

The response time of the proposed design is evaluated while scaling the circuit size from 50 cells up to 1000. The circuit is setup as follows: every 50 cells are placed in a single cluster. All cell inputs are set to 0.6v, except the one which is connected to a pulse signal with a period of 300 ns. When two clusters are placed in the same testing environment, only one cell is left to be connected to a pulse signal whereas the rest are connected to a fixed DC voltage, 0.6v. 
The experiment is done in iterative form. In each iteration 50 more cells are added to the testing environment until 1000 cells is reached. For the purpose of comparison, all cells are loaded with the same capacitive load, 1 pF, as in ~\cite{choi1993high}. It is found that the response time of the proposed WTA circuit is 6.36 ns and it is independent of the number of cells. This can be mainly attributed to driving the load with buffers which allows for the circuit to have a stable output and high drive strength. Additionally, the normalized exponential relationship between the input and output makes most of the bias current, $I_c$, directed to the winning cell while the losing cells would have almost negligible current. 
~\fig{accur}-(c) shows the response time when simulating two cells loaded with 1 pF. One of the cells is connected to a fixed input voltage VDD/2 while the other input is connected to a pulse signal of width 150 ns. 

\begin{table}[ht]
\caption{Transistor sizes for one cell in the proposed WTA circuit (with the excitatory feedback).}
\label{sizes}
\begin{center}
\begin{tabular}{|a|c|c|c|c|c|}
  \hline              
\textbf{Transistors} & \textbf{$M_1$} & \textbf{$M_2$} & \textbf{$M_3$} & \textbf{$M_4$} & \textbf{$M_f$}\\ \hline
\textbf{W/L [nm/nm]} & 350/60 & 500/60 & 500/60 & 500/60 & 120/60\\ \hline
\end{tabular}
\end{center}
\end{table}

The list of the circuit parameters and how to set them up is presented below:
\begin{itemize}
\item $I_c$ controls the network precision and is set based on the desired accuracy (the minimum voltage between two competitors to set one of them as winner). In case of $I_m$, it is chosen based on the desired number of winners, $\delta$, and according to \eq{Ic_eq}.
\begin{equation}
\frac{I_c}{2\delta} \le I_{m} < \frac{I_c}{\delta}
\label{Ic_eq}
\end{equation}

\item The input transistors are sized to be minimum when no local excitatory feedback is used to optimize the circuit for the area. In the presence of feedback, the input and feedback transistors are sized based on the desired width of the hysteresis curve. In case of the mirroring transistors, high dimension transistors are recommended since the device's length influences the ability to eliminate the channel modulation effect.  
\end{itemize}


\section{WTA Applications}
The WTA circuits can be exploited in a wide range of applications such as spatial filtering, object tracking, etc~\cite{fish2001cmos}. 

\textbf{Application 1:} The WTA circuit has a unique ability in removing images' background and binarizing the images. The latter can be conducted by performing a hardmax function which involves zeroing all the pixels with low intensity, i.e. below a predefined threshold, and set the rest to maximum, which is 1.2v in this work, as shown in \fig{catii}. Binarizing the images is a useful process, especially for bio-inspired algorithms that work on the basis of sparse representation such as hierarchical temporal memory (HTM)~\cite{hawkins2010hierarchical, zyarah2015reconfigurable, zyarah2018neuromorphic}. However, this kind of task can not be achieved with a WTA circuit unless it has the capability to enable more than one winner.  

\begin{figure} [h!tb]
\begin{center}
\includegraphics[width = 0.4 \textwidth]{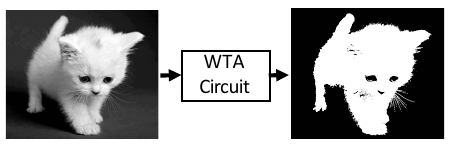}
\caption{Binarizing grayscale image using a WTA circuit.}
\label{catii}
\end{center}
\end{figure}

\begin{figure} [h!tb]
\begin{center}
\includegraphics[width = 0.45 \textwidth]{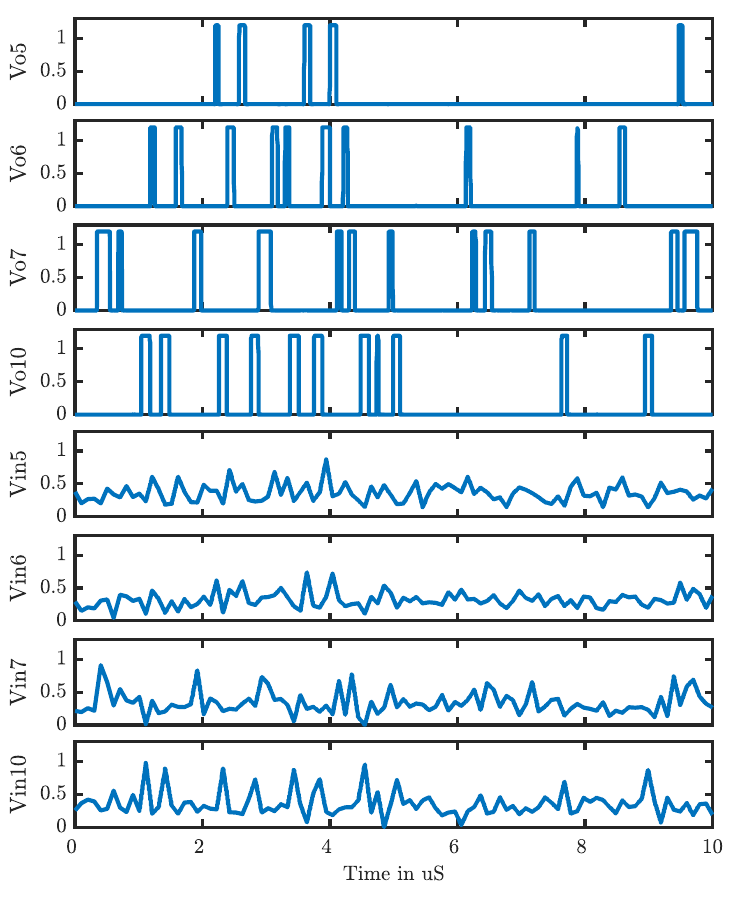}
\caption{Simulation results of four randomly picked cells from the proposed WTA circuit while identifying the winning class in a densely connected convolution network. For the shown waveforms, the expected output label is: [x-7-7-x-7-x-10-6-x-10-...], where x indicates other classes (not shown here). } 
\label{waves}
\end{center}
\end{figure}

\textbf{Application 2:} In this work, the proposed WTA circuit is used to perform the softmax operation for a densely connected convolution network~\cite{huang2017densely}. The network is composed of five main layers. The first of which is the convolution layer which performs feature extraction on the input image. This is followed by a dense block that concatenates features and propagates these features such that catastrophic forgetting does not occur. Down sampling is then conducted in the pooling layer. The output of the pooling layer is relayed to a fully-connected and softmax layer to predict the class label. The network is trained and tested on image classification task using CIFAR-10 dataset~\cite{krizhevsky2009learning} with 90\% testing accuracy. For the same testing task, the output of the network fetched to the softmax layer is normalized to be between 0-1 and exported to Cadence as an analog signal. In Cadence, the data are imported by using the piece-wise source and presented to the WTA circuit to identify the winning class at each point in time. ~\fig{waves} depicts the input signals to four randomly picked cells and their corresponding digital outputs. Each input signal is formed by a piece-wise operation connecting the discrete testing points over 100 consecutive samples. The output of the softmax is compared to the labels associated with each input sample and a perfect match is achieved. However, the resulting digital output signal does not have a unified pulse period, i.e. some may have longer pulse than others. 
This is due to the irregular nature of the piece-wise signals and can be modulated with predefined constraints. 

\begin{figure} [h!tb]
\begin{center}
\includegraphics[width = 0.45 \textwidth]{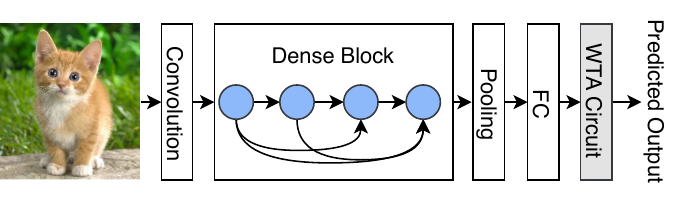}
\caption{Densely connected convolution network with one dense block.}
\label{catii}
\end{center}
\end{figure}

\section{Simulation Results} 

\begin{table*}[h!tb]
\caption{List of the WTA circuits that include both analog and digital implementations. Data is obtained from \cite{yoon2015vlsi,choi1993high,starzyk1996voltage,fish2005high,prommee2011cmos,rahman2009high,kulej2017sub}. }
\label{HardwareAnalysis}
\setlength\tabcolsep{4 pt}
\begin{center}
\begin{threeparttable}
\begin{tabular}{|c|c|c|c|c|c|c|c|c|c|}
\hline                      
\rowcolor{Gray} \textbf{Parameter} & \textbf{\cite{choi1993high}} & \textbf{\cite{starzyk1996voltage}} & \textbf{\cite{fish2005high}} & \textbf{SDWTA~\cite{yoon2015vlsi}} & \textbf{\cite{prommee2011cmos}} & \textbf{\cite{rahman2009high}} & \textbf{VF-WTA\cite{kulej2017sub}} & \textbf{This work} \\ \hline
Design & Analog & Analog  & Analog & Digital & Analog & Analog & Analog   & Analog  \\ \hline
Input&Voltage & Voltage &Current & Voltage & Voltage & Voltage&  Voltage  & Voltage\\ \hline
Output &Voltage & Voltage &Voltage &Voltage & Voltage &Voltage & Voltage  & Voltage\\ \hline
Voltage Supply & 5v & 5v & 3.3v &1.2v & $\pm$1.25 & 1.8v & 0.5v-0.3v & 1.2v \\ \hline
Resolution  & - & 50 mv & 1.8 nA & 1/64k & - & 0.5 mV& - & 4 mv \\ \hline
Latency  & 60 ns & 12 ns\tnotex{tnote:robots-a2} & 34 ns & 16.5 ns  & - & 30 ns  & - & 17.6 ns \\ \hline
No. of Inputs\tnotex{tnote:robots-a3} & 1000 & 10 & 8 & 1024 & 2 & 4 & 2 & 1000 \\ \hline
Power dissipation\tnotex{tnote:robots-a1} & 120 $\mu$W & - & 22.5 $\mu$W & -  & 0.31 mW &- &1.75 $\mu$W - 42 nW & 34.9 nW \\ \hline
Technology node & Orbit 2 $\mu$m & 2 $\mu$m & 0.35 $\mu$m &IBM 130 nm & TSMC 0.25 $\mu$m& GPDK 90 nm &0.35 $\mu$m & IBM 65 nm   \\ \hline
  \end{tabular}
      \begin{tablenotes}
\item\label{tnote:robots-a1} Power dissipation per cell with load of 1 pF. 
\item\label{tnote:robots-a2} Nominal delay time (the authors did not report the worst case delay).
 \item\label{tnote:robots-a3} Simulated number of inputs.
    \end{tablenotes}
\end{threeparttable}
\end{center}
\end{table*}

In order to evaluate the proposed circuit, the commonly used metrics, speed, resolution, and power consumption~\cite{rahman2009high}, are considered.~\tb{HardwareAnalysis}\footnote{This table is included to provide a high level overview about the previous work. In fact, the characteristics of the WTA is highly affected by the technology node and the specific optimization constraints the circuit is design for.} summarizes the main characteristics of the proposed circuit and the previous ones. It can be noticed that our design combines most of the desired features needed in a WTA circuit including scalability, high speed, and low power consumption. Regarding the scalability, as in~\cite{choi1993high,yoon2015vlsi}, the circuit can be scaled up to 1000 cells without any noticeable issue. In case of worst case delay, we achieved a latency comparable to that achieved in the digital WTA circuit by~\cite{yoon2015vlsi} and less than that in~\cite{starzyk1996voltage} for the nominal operation in which our design has a latency of 10.4 ns. The proposed design also offers low power consumption achieved via implementing the softmax equation which limits the activity of the circuit. The power consumption is even less than that achieved in bulk-driven WTA in~\cite{kulej2017sub}. 


The power consumption of the proposed WTA circuit is measured using Cadence-ADE with IBM 65nm technology node. A WTA circuit with 1000 cells is tested. The circuit is implemented in form of clusters where every 50 cells are assigned to one cluster. Each cluster has its own dedicated $I_m$ source while all the 1000 cells have one shared bias current, $I_c$. In a given cluster, all the cell inputs except one, are connected to DC voltage source of 0.7v, while the left cell from each cluster is connected to an AC signal (randomly picked, either a sine, pulse, or triangle). These signals have a voltage ranged between 0-1.2v, frequency less than 0.5 MHz, and random delay to avoid the perfect overlap. Each cell output is loaded with a capacitive load of 1 pF. For the given setup, the measured power is estimated to be 34.89 $\mu W$ for the entire architecture. Interestingly, the power consumption is not effected much by the scalability due to the fact that the cells which have small voltages consume negligible amounts of power. This occurs due to the nature of the circuit which performs the softmax operation leading to suppression of the amount of current in the losing cells to negligible amounts and directing most of the bias current to the winning ones. To investigate this further, we performed the corner analysis.

\begin{figure} [h!tb]
\begin{center}
\includegraphics[width = 0.35 \textwidth]{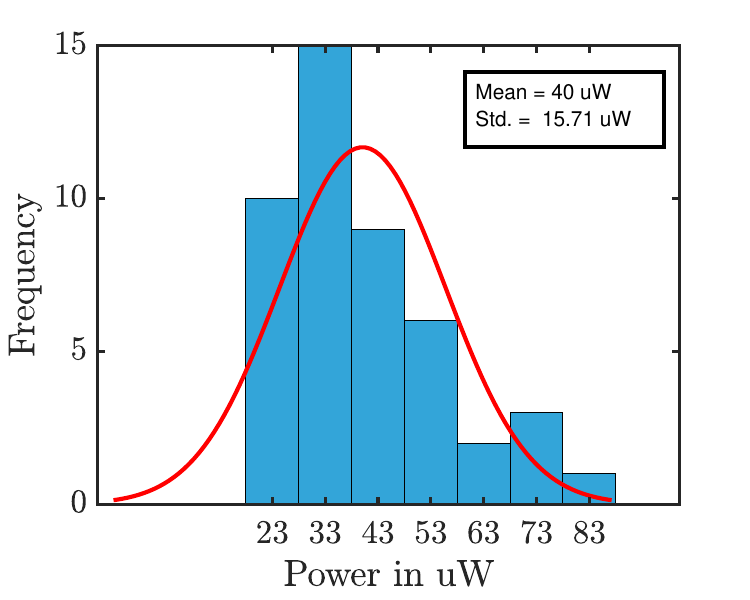}
\caption{Power consumption histogram of the proposed WTA circuit with 1000 cells while running corner analysis.}
\label{power_corner}
\end{center}
\end{figure}

\begin{figure} [h!tb]
\begin{center}
\includegraphics[width = 0.35 \textwidth]{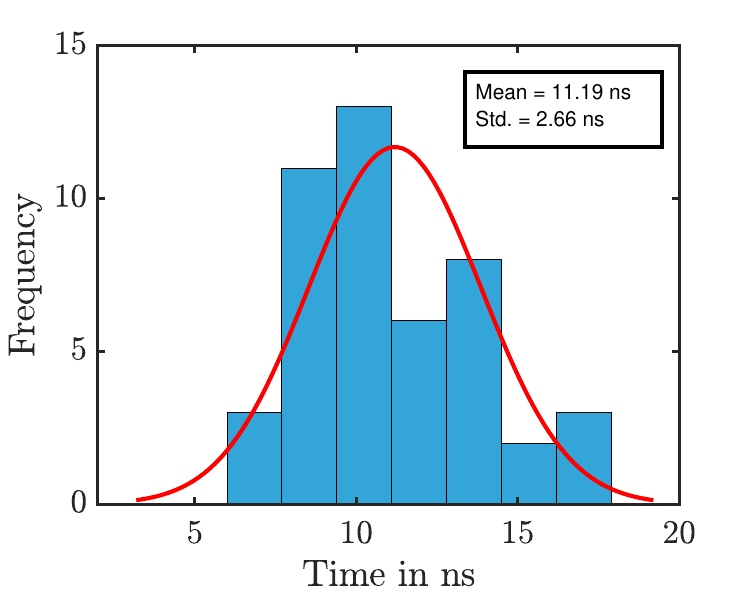}
\caption{Latency histogram of the proposed WTA circuit with 1000 cells while running corner analysis.}
\label{hist_delay}
\end{center}
\end{figure}

The corner analysis is used to evaluate the proposed WTA circuit performance under different conditions. The design is tested for fabrication process (FF, FS, SF, SS, TT), voltage supply (1.32v, 1.2v, 1.02v), and ambient temperature (\SI{100}{\degreeCelsius}, \SI{27}{\degreeCelsius}, \SI{0}{\degreeCelsius}) (PVT) variations.~\fig{power_corner} illustrates the histogram plot of the power consumption with a mean and standard deviation of 40 $\mu$W and 15.71 $\mu$W, respectively. It can be noticed that having 10\% variation in the voltage supply rail and sizing the input transistors to be small together with their operation in the subthreshold region makes the circuit sensitive to PVT variations. One possible solution to reduce this sensitivity involves increasing device sizes and reducing the voltage supply rail variation to be 5\% or less. During the corner analysis for power consumption, the worst case scenario is identified in FF, 1.32v, \SI{100}{\degreeCelsius} in which the power is measured to be 84.83 $\mu$W.  Regarding the circuit latency, the corner analysis result is shown in~\fig{hist_delay}, with mean of 11.19 ns and standard deviation of 2.66 ns. The results shows that the proposed circuit nominal latency is 10.4ns. The latency for the worst case scenario is achieved in FS, 1.08v, \SI{100}{\degreeCelsius}, which is measured to be 17.61 ns. 

\section{Conclusions}
This article presents a voltage mode WTA circuit optimized for power and performance and designed in 65nm technology node. The utility of the circuit is demonstrated on spatial filtering and classification applications. The proposed circuit is well suited for neuromorphic chips especially that require WTA circuit such as spiking neural networks and hierarchical temporal memory.

\bibliographystyle{ACM-Reference-Format}
\bibliography{Refer}

\end{document}